\documentclass{article}
\usepackage{ijcai23}

\usepackage{times}
\usepackage{soul}
\usepackage{url}
\usepackage[hidelinks]{hyperref}
\usepackage[utf8]{inputenc}
\usepackage[small]{caption}
\usepackage{graphicx}
\usepackage{amsmath}
\usepackage{amsthm}
\usepackage{booktabs}
\usepackage{color}
\usepackage{algorithm}
\usepackage{algorithmic}
\usepackage[switch]{lineno}
\usepackage{amssymb}
\usepackage{setspace}

\usepackage{multirow}
\usepackage{bbold}

\title{MILD: Modeling the Instance Learning Dynamics for Learning with Noisy Labels}

\author{
Chuanyang Hu$^1$\footnote{Code is available at \url{https://github.com/Hcyang-NULL/MILD}}
\and
Shipeng Yan$^1$\and
Zhitong Gao$^1$\And
Xuming He$^{1,2}$
\affiliations
$^1$ShanghaiTech University\\
$^2$Shanghai Engineering Research Center of Intelligent Vision and Imaging
\emails
\{huchy3, yanshp, gaozht, hexm\}@shanghaitech.edu.cn
}

\begin{document}

\maketitle

\begin{abstract}
Despite deep learning has achieved great success, it often relies on a large amount of training data with accurate labels, which are expensive and time-consuming to collect.
A prominent direction to reduce the cost is to learn with noisy labels, which are ubiquitous in the real-world applications.
A critical challenge for such a learning task is to reduce the effect of network memorization on the falsely-labeled data. 
In this work, we propose an iterative selection approach based on the Weibull mixture model, which identifies clean data by considering the overall learning dynamics of each data instance.
In contrast to the previous small-loss heuristics, we leverage the observation that deep network is easy to memorize and hard to forget clean data. In particular, we measure the difficulty of \textit{memorization} and \textit{forgetting} for each instance via the transition times between being misclassified and being memorized in training, and integrate them into a novel metric for selection.
Based on the proposed metric, we retain a subset of identified clean data and repeat the selection procedure to iteratively refine the clean subset, which is finally used for model training.
To validate our method, we perform extensive experiments on synthetic noisy datasets and real-world web data, and our strategy outperforms existing noisy-label learning methods.
\end{abstract}

\section{Introduction}
Deep neural networks has achieved impressive success in a wide spectrum of machine learning tasks. However, they usually rely on a large amount of data with accurate supervision, which is expensive and time-consuming to obtain in practice.
In real-world applications, a common strategy for collecting large-scale data is to use online queries~\cite{blum2003noise} and/or crowdsourcing~\cite{yan2014learning}, which are typically cheap but the resulting annotations are noisy.
Nonetheless, recent studies~\cite{arpit2017closer} demonstrate that the capacity of deep networks is sufficient to fit even randomly generated labels, which hinders the generalization when the deep networks learn with noisy labels.
Therefore, it is essential to develop robust training methods with noisy labels for deep neural networks, which can benefit a broad scope of applications such as medical diagnosis~\cite{li2021superpixel} and autonomous driving systems~\cite{feng2020deep}.

In order to alleviate the impact of noisy data, it is critical to identify the clean subset of data, also referred to as sample selection~\cite{song2022learning}.
By removing mislabeled data as much as possible, we are able to reduce the impact of mislabeled data and robustify the learning process.
During this process, perhaps the most important aspect of sample selection is the validity of the data selection metric.
There are some existing works exploring the design of metric.
Small-loss metric is the most popular selection approach, which considers samples with small loss values to be clean. 
This is based on the observation that deep networks learn easy patterns first, and then they overfit to the noisy pattern~\cite{arpit2017closer}.
Recently, FINE~\cite{FINE} utilizes the principal component of feature representation and split the data according to the projection to the principal component.
However, those criteria depends on the current training state of models for removing corrupted data and are often unstable.

To tackle the above-mentioned issues, we propose a novel sample selection framework for learning from noisy labels. 
In contrast to prior works that simply utilize current state of a trained model, we adopt a novel perspective that exploits the overall learning dynamic to select clean training data.
Our main idea is to design a new metric to differentiate between the clean and corrupted data based on the observation that the network is easy to memorize and hard to forget clean samples. To this end, we introduce memorization and forgetting measurements to represent the overall learning dynamic for each instance, which are integrated into our metric. Moreover, we build a mixture model of the metric distribution to determine the selection threshold, and perform multiple training rounds to iteratively refine the selected clean data.

Specifically, 
at each round, we distinguish examples based on the metric which takes into account the \textit{memorization} and \textit{forgetting} property of each data during the learning process.
To this end, we record the prediction history of each instance during the training as a sequence, and 
measure the difficulty of memorization by the transition times from being misclassified to memorized and the difficulty of forgetting by the transition times from being memorized to misclassified.
We then develop a selection metric by combining the memorization and forgetting measurements, as they play a complementary role in describing the learning dynamics.
To perform selection, we adopt a Weibull mixture model to fit the metric distribution and determine the selection threshold. 
Finally, to take advantage of the falsely-labeled data, we further extend our sample selection approach by incorporating semi-supervised techniques in the model training.


To validate the effectiveness of our proposed approach, we perform extensive experiments on CIFAR-10~\cite{cifar-dataset}, CIFAR-100~\cite{cifar-dataset}, Mini-WebVision~\cite{webvision} and Mini-ImageNet~\cite{mini-imagenet} datasets under different noise ratios and types.
The experimental results show that our method consistently outperforms other methods, thereby demonstrating its superiority.
Besides, we also perform detailed analysis of our approach to provide more insights on the new metric.

Our main contributions can be summarized as follows,
\begin{itemize}
    \item We propose a novel selection metric by taking advantage of instance learning dynamics to discriminate clean and corrupted data. Our strategy incorporates the learning dynamics from both instance memorization and forgetting perspective.
    \item We develop a novel sample-selection approach based on a Weibull mixture model on the metric distribution, which refines the clean data selection via an iterative training strategy.
    \item Our approach achieves the state of the art on five popular noisy image classification benchmarks. We also show that our method can be effectively combined with multiple semi-supervised methods.
\end{itemize}

\section{Related Works}
In this section, we briefly review existing relevant works on learning with noisy labels.
Noisy data interfere with the learning of clean data~\cite{arpit2017closer}.
To reduce the influence, the approach can be mainly summarized as follows: sample-selection methods, semi-supervised based methods and other noisy label approaches.

\begin{figure*}[!h] 
    \centering 
    \includegraphics[width=\textwidth]{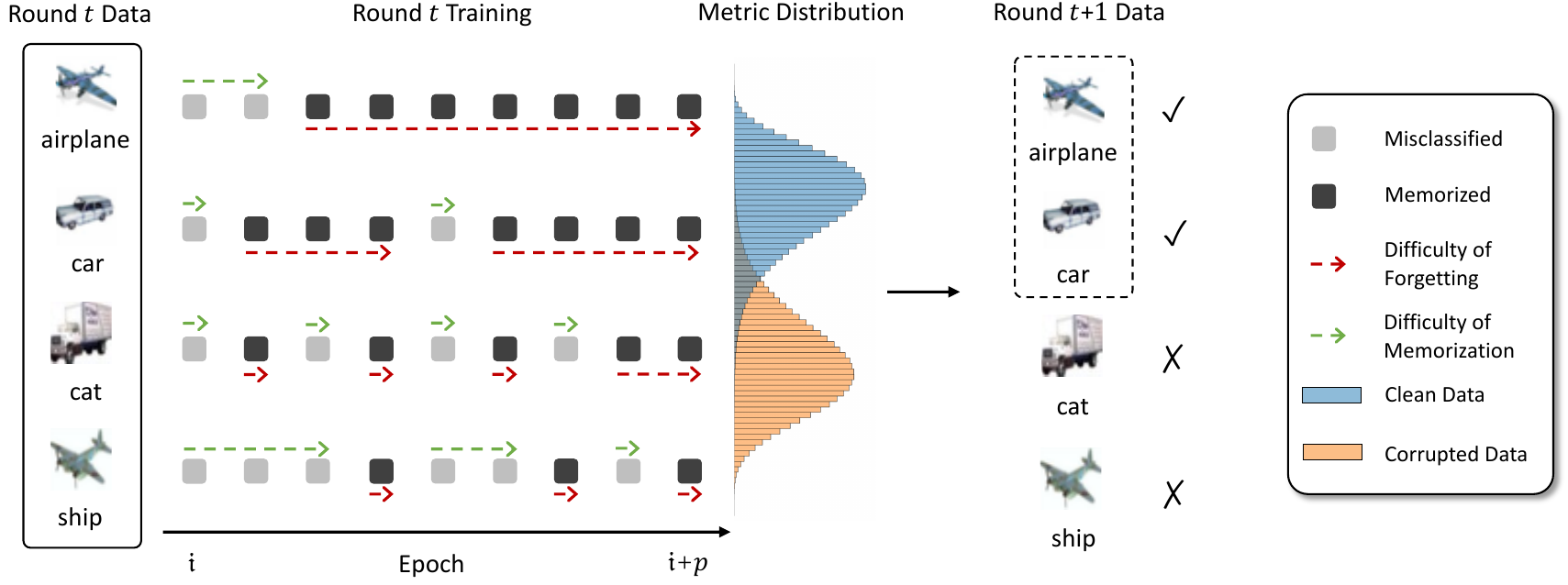} 
    \caption{The overview of each sample selection round. 
    The input dataset at left is noisy like the truck image with cat label.
    To model learning dynamic for each instance, we record the learning status at each epoch as a sequence, where gray and black squares indicate misclassified and memorized.
    The $p$ is the number of epochs for current round.
    Note that the total length of green and red arrows indicate the difficulties of memorization and forgetting.
    Then we compute the selection metric based on the difficulties.
    Finally, we identify and select the clean data, and then use them for subsequent round.
} 
    \label{method_fig} 
\end{figure*}

\paragraph{Sample-selection methods}
To avoid the influence of data noises, sample selection approaches~\cite{song2022learning} to detect and filter falsely-labeled data are proposed.
It is critical to choose an appropriate criterion to separate clean and corrupted samples when deciding on a sample selection procedure.
Small-loss~\cite{arpit2017closer} is the most widely utilized criterion.
The empirical findings of study~\cite{arpit2017closer} indicate that deep neural networks acquire progressively more sophisticated hypotheses and learn the simple patterns of clean data prior to fitting noisy data, supporting the reasonableness of small-loss criterion.
Few works~\cite{nguyen2019self,zhou2020robust} adopt the exponential moving average of loss to select clean data.
FINE~\cite{FINE} filters noisy instances by their eigenvectors which computes the alignment of image feature and the principal component of gram matrix in the learned representation space.
\cite{zhu2021understanding} uses the number of projected gradient steps to attack a data point to assess a data point's distance to its class boundary, which can then be used to distinguish between clean and corrupted data.

In addition to the selection criterions, there are also works~\cite{song2022learning} that employ several networks to collaborate with one another or multi-round learning to iteratively enhance the selected dataset.
MentorNet~\cite{jiang2018mentornet} introduce mentor network to supervise the training of a student network, where MentorNet provides a curriculum for StudentNet's training to concentrate on the clean samples.
JoCoR~\cite{jocor} employs two networks to make predictions on the same mini-batch data and improves the agreement of two network via computing a joint loss with Co-Regularization for each training instance.
In Co-teaching~\cite{han2018co}, each network selects its small-loss instances and feeds them to its peer network, which can filter various sorts of noise errors because of the distinct each network's learning ability.
Multi-round learning refines the selected subset repeatedly, which keeps improving the quality of selected set as the number of rounds increase.
For multi-round learning, INCV~\cite{chen2019understanding} introduces cross-validation to randomly partition noisy training data in order to determine clean data for the subsequent round training. 
It also employs Co-teaching strategy to exploit the selected data to train deep networks.
ITLM~\cite{shen2019learning} iteratively minimize the trimmed loss by repeating multiple rounds.
For each round, it selects small loss samples to build a more clean dataset and then retrain the deep network using selected data.
In order to gain from falsely-labeled data, semi-supervised techniques are also explored to combine with sample selection to further improve performance~\cite{li2020dividemix,moit}.
In contrast to previous approaches simply considering the present state of the model, we distinguish the clean and corrupted data based on their learning dynamics and then do multi-round learning.

\paragraph{Semi-supervised learning}
There are some works \cite{li2020dividemix,moit,song2019selfie,zhou2020robust} trying to leverage the falsely-labeled data in an semi-supervised(SSL) manner.
DivideMix~\cite{li2020dividemix} first perform sample selection to eliminate the samples that are more likely to be corrupted, and utilize the noisy samples as unlabeled data to regularize training in a SSL manner with MixMatch~\cite{berthelot2019mixmatch}.
MOiT~\cite{moit} introduces interpolated supervised contrastive loss with MixUp~\cite{zhang2017mixup} to interpolate data points in order to enhance model performance.
ScanMix~\cite{zhang2017mixup} develops from DivideMix and introduce  Expectation-Maximization(EM) based semantic clustering to improve the robustness to noisy label.
SELFIE~\cite{song2019selfie} corrects refurbishable example, which can be corrected with high precision.
Concretely, it considers the example with consistent label predictions as refurbishable examples.
In this study, we also demonstrate that our method can be used effectively with a variety of semi-supervised methodologies.

\paragraph{Other noisy label methods}
~\cite{ghosh2017robust} demonstrates that risk minimization with mean absolute error(MAE) can be more tolerant of noisy labels, but is not appropriate for complicated data.
Generalized cross entropy(GCE)~\cite{zhang2018generalized} is a more general kind of noise-robust loss that combines the benefits of categorical cross entropy loss(CCE) and MAE.
APL~\cite{ma2020normalized} further combines existing robust loss functions that mutually boost each other.
Other works utilize implicit/explicit regularization~\cite{elr} or transition matrix learning\cite{CAL} to improve the robustness of the model.
MixUp~\cite{zhang2017mixup} training strategy that greatly prevents memorization to noise.
Self-training~\cite{arazo2019unsupervised} has been employed to generate pseudo label to correct the falsely-labeled data.

\section{Method}
In this section, we present our approach to the robust learning of image classification tasks, aiming to improve the generalization of classification model.
To this end, we propose a sample selection approach leveraging the instance learning dynamics to separate the clean and falsely-labeled data.
Below we first present the overall sample selection framework in Sec.~\ref{subsec:sample_selection}.
Then we introduce the design of proposed selection metric using learning dynamics in Sec.~\ref{subsec:metric} and the selection strategy based on a Weibull mixture model in Sec.~\ref{subsec:wmm}.
Finally, we present the exploitation of corrupted data in Sec.~\ref{subsec:semisup}.

\subsection{Multi-round Sample Selection Framework}\label{subsec:sample_selection}
The noisy training set $\mathcal{D}= \{(x_i, y_i)\}_{i=1}^{|\mathcal{D}|}$ where $x_i$, $y_i$ is the i-th image and its corresponding label.
Note that there are some images that are falsely labeled.
During training, our sample selection will repeat for multiple rounds.
We adopt the categorical cross entropy loss on current training set $\mathcal{D}$ to update the model.
Algorithm~\ref{algo:sample_select} describes the selection procedure of one round.
As training progress, we continually update selection metric values.
At the end of each round, we apply a Weibull mixture model to identify and choose the clean data as the training set $\mathcal{D^{'}}$ for next round.

\subsection{Selection Metric Design} 

\paragraph{Model the instance learning dynamics}
In this work, we distinguish the corrupted and clean data via their learning dynamics.
First, we represent the instance learning dynamic with instance prediction sequence $S_i$ for each instance $x_i$, which records the instance learning status for each epoch.
Concretely, given an image $x_i$, the model outputs the prediction label $\hat{y}=\arg \max_{c} p(y=c|x)$.
The example $x_i$ is deemed to be \textit{memorized} if the prediction label $\hat{y}$ equals the dataset label $y$. If not, it is referred to as \textit{misclassified}.
Based on the learning status, we can obtain the k-th element of instance prediction sequence $S^k_i=\mathbb{1}_{\hat{y}_i=y_i}$, which corresponds to the epoch $k$.
Besides, the indicator function $\mathbb{1}_{\hat{y}_i=y_i}$ is 1 when $\hat{y}_i=y_i$, otherwise 0. 
Note that we record instance's learning status before updating the model with gradient.
It is worth noting that as learning progresses, the proportion of memorized examples keep increasing.


\paragraph{Selection metric}\label{subsec:metric}
We can further summarize the instance learning dynamics to distinguish between clean and corrupted data.
Note that ~\cite{arpit2017closer} finds that the model often first learns the patterns of clean data and then overfits to corrupted data.
Besides, ~\cite{toneva2018empirical} observes that the corrupted data is more easy to forget compared to clean data.
Motivated by these observations, we can hypothesis that clean samples are both easy to learn and hard to forget.
Furthermore, we estimate the difficulty of memorization and forgetting based on the instance prediction sequence $S_i$.
Therefore, we propose a novel selection metric by combining them as we believe that memorization and forgetting are complementary to separate clean and corrupted data.


The intuitive solution to estimate the difficulty of memorization is to measure the average transition times from misclassified to memorized.
Similarly, the difficulty of forgetting can be estimated via the average transition times from memorized to misclassified.
An instance prediction sequence $S_i^k$ may have several memorization and forgetting events during learning, as shown in 
Fig.~\ref{method_fig}.
Therefore, we split the instance sequence $S_i$ into continuous segments $G_i=\{g_{i,1}^u, \cdots , g_{i,n_i^u}^u\} \cup \{g_{i,1}^l, \cdots, g_{i,n_i^l}^l\}$ with same indicator value $S^k_i$ in segment,
where the segments $g_j$ are disjoint. 
It is worth noting that $g^u_{i,j}$ means the segment where the instances have not been memorized.
Similarly, $g^l_{i,j}$ refers to the segment where the instances are memorized by the model.
$n_i^u$ and $n_i^l$ are the number of misclassified segments $g^u_i$ and memorized segments $g^l_i$, respectively.
Therefore, given instance prediction sequence $S_i$, we define the difficulty of memorization $M_i$ and forgetting $F_i$ as follows 


\begin{equation}
    \begin{aligned}
        M_i = \frac{\sum_{j=0}^{n_i^u}|g_j^u|}{n_i^u}, \quad F_i= \frac{\sum_{j=0}^{n_i^l}|g_j^l|}{n_i^l} \\
    \end{aligned}
    \label{eq:mem_for_def}
\end{equation}
where $|\cdot|$ represents the segment length.
Considering that clean instances typically have higher $F_i$ and smaller $M_i$, we create the selection metric by combining the memorization and forgetting in the way described below
\begin{equation}
    \begin{aligned}
        C^F_i = M_i - \lambda F_i
    \end{aligned}
    \label{mf_def}
\end{equation}
where $\lambda$ is the coefficient to adjust the influence of  memorization and forgetting. Typically, we take $\lambda=1$ in experiment.
Moreover, our metric can be simplified by omitting the number of segments from equation~\ref{eq:mem_for_def}.
Empirically we find that the simplified version is very close to the full version $C^F$.
Therefore, our metric after simplification is stated as follows
\begin{equation}
    \begin{aligned}
        C_i = \tilde{M}_i - \lambda \tilde{F}_i  = n_i^u M_i - \lambda n_i^l F_i = \sum_{j=0}^{n^u_i} |g_j^u| - \lambda \sum_{j=0}^{n_i^l} |g_j^l|
    \end{aligned}
    \label{final_metric}
\end{equation}
where $\tilde{M}_i$, $\tilde{F}_i$ are to measure the difficulty of memorization and forgetting when ignoring the number of segments.

\subsection{Selection Strategy}\label{subsec:wmm}

Based on empirical observation, we adopt the Weibull distribution to represent the distribution of our metric scores on the data with clean or noisy labels. The probability density function of Weibull distribution is defined as follows

\begin{equation}
    \begin{aligned}
        \phi(x|\alpha, \beta) = \frac{\beta}{\alpha}(\frac{x}{\alpha})^{\beta-1}e^{-(x / \alpha)^\beta}
    \end{aligned}
    \label{Weibull_pdf}
\end{equation}
where $x$ is the metric distribution, $\alpha$ is the scale parameter and $\beta$ is the shape parameter. 

Given the metric distribution $F_C$, we use the Weibull mixture model which contains two components to fit the metric distributions of clean and falsely-labeled data.

\begin{equation}
    \begin{aligned}
        P(x|\theta) = k_1 * \phi_c(x|\alpha_1, \beta_1) + k_2 * \phi_f(x|\alpha_2, \beta_2)
    \end{aligned}
    \label{wmm}
\end{equation}
where $\theta = (\alpha_1, \beta_1, \alpha_2, \beta_2)$, $k_1$ and $k_2$ are the coefficients of each component of the mixture model which subjects to $k_1 + k_2 = 1$, $\phi_c$ and $\phi_f$ represent the distribution of clean and falsely-labeled data respectively. According to the definition of our metric, the component of the Weibull mixture model corresponding to the falsely-labeled data is typically has a larger mean value, as the falsely-labeled data is difficult to memorize and easy to forget, resulting in a relatively large metric value. Consequently, we are able to determine which component corresponds to the clean or falsely-labeled data. 

Finally, we choose the threshold $\tau = \alpha_2$ and select all the training data instances with a score smaller than $\tau$ for the next round. The reason for not choosing the position where the two components have the same probability as the threshold is because the performance will deteriorate due to the sharp drop of recall in multiple 
rounds. Empirically, we found choosing the threshold $\tau$ can be beneficial to maintain a high recall while also reducing the computational complexity.

\begin{table*}[t]
    \renewcommand\arraystretch{1.5}
    \centering
    {\resizebox{\linewidth}{!}{
    \begin{tabular}{c|cccc|cccc} 
    \toprule
    Dataset      & \multicolumn{4}{c|}{CIFAR-10}                                     & \multicolumn{4}{c}{CIFAR-100}                                      \\ 
    \midrule
    Noisy Type   & \multicolumn{3}{c}{Symmetry}                     & Asymmetry      & \multicolumn{3}{c}{Symmetry}                     & Asymmetry       \\ 
    \midrule
    Noise Ratio  & 20             & 50             & 80             & 40             & 20             & 50             & 80             & 40              \\ 
    \midrule
    Standard     & 87.0 $\pm$ 0.1  & 78.2 $\pm$ 0.8  & 53.8 $\pm$ 1.0  & 85.0 $\pm$ 0.0  & 58.7 $\pm$ 0.3  & 42.5 $\pm$ 0.3  & 18.1 $\pm$ 0.8  & 42.7 $\pm$ 0.6   \\
    Bootstrap\cite{bootstrap}    & 86.2 $\pm$ 0.2 & - & 54.1 $\pm$ 1.3 & 81.2 $\pm$ 1.5 & 58.3 $\pm$ 0.2 & - & 21.6 $\pm$ 1.0 & 45.1 $\pm$ 0.6  \\
    Forward\cite{forward}      & 88.0 $\pm$ 0.4 & - & 54.6 $\pm$ 0.4 & 83.6 $\pm$ 0.6 & 39.2 $\pm$ 2.6 & - & 9.0 $\pm$ 0.6  & 34.4 $\pm$ 1.9  \\
    Co-teaching\cite{coteaching}  & 89.3 $\pm$ 0.3 & 83.3 $\pm$ 0.6 & 66.3 $\pm$ 1.5 & 88.4 $\pm$ 2.8 & 63.4 $\pm$ 0.0 & 49.1 $\pm$ 0.4 & 20.5 $\pm$ 1.3 & 47.7 $\pm$ 1.2  \\
    Co-teaching+\cite{coteaching+} & 89.1 $\pm$ 0.5 & 84.9 $\pm$ 0.4 & 63.8 $\pm$ 2.3 & 86.5 $\pm$ 1.2 & 59.2 $\pm$ 0.4 & 47.1 $\pm$ 0.3 & 20.2 $\pm$ 0.9 & 44.7 $\pm$ 0.6  \\
    TopoFilter\cite{topfilter}   & 90.4 $\pm$ 0.2 & 86.8 $\pm$ 0.3 & 46.8 $\pm$ 1.0 & 87.5 $\pm$ 0.4 & 66.9 $\pm$ 0.4 & 53.4 $\pm$ 1.8 & 18.3 $\pm$ 1.7 & 56.6 $\pm$ 0.5  \\
    CRUST\cite{crust}        & 89.4 $\pm$ 0.2 & 87.0 $\pm$ 0.1 & 64.8 $\pm$ 1.5 & 82.4 $\pm$ 0.0 & 69.3 $\pm$ 0.2 & 62.3 $\pm$ 0.2 & 21.7 $\pm$ 0.7 & 56.1 $\pm$ 0.5  \\
    JoCoR\cite{jocor}$^\dag$        & 90.2 $\pm$ 0.4 & 47.9 $\pm$ 2.2 & 24.1 $\pm$ 1.9 & 74.1 $\pm$ 0.3 & 64.8 $\pm$ 1.2 & 51.2 $\pm$ 1.4 & 9.2 $\pm$ 2.6 & 42.1 $\pm$ 0.8 \\
    FINE\cite{FINE}         & 91.0 $\pm$ 0.1 & 87.3 $\pm$ 0.2 & 69.4 $\pm$ 1.1 & 89.5 $\pm$ 0.1 & 70.3 $\pm$ 0.2 & 64.2 $\pm$ 0.5 & 25.6 $\pm$ 1.2 & 61.7 $\pm$ 1.0  \\
    MILD & \textbf{93.0 $\pm$ 0.2} & \textbf{88.7 $\pm$ 0.2} & \textbf{79.1 $\pm$ 0.5} & \textbf{89.8 $\pm$ 0.3} & \textbf{74.2 $\pm$ 0.2} & \textbf{67.3 $\pm$ 0.3} & \textbf{36.0 $\pm$ 0.3} & \textbf{69.9 $\pm$ 0.6} \\
    \bottomrule
    \end{tabular}}}
    \caption{Test accuracy (\%) on CIFAR-10 and CIFAR-100 under Resnet34 backbone with different noise ratios and types. The average accuracies and standard deviations over three trials are reported. $\dag$ means the results are reproduced.}
    \label{cifar-results}
\end{table*}

\begin{algorithm}[t]
	\renewcommand{\algorithmicrequire}{\textbf{Input:}}
	\renewcommand{\algorithmicensure}{\textbf{Output:}}
	\caption{Pseudocode of Sample Selection}
	\label{algo:sample_select}
	\begin{algorithmic}[1]
	    \REQUIRE Current training set $D$, Model $M$, Number of Epoch $E$
	    \STATE Initialization: $e \leftarrow 0$
	    \FOR{$e = 0; e < E; e$++}
            \FOR{Batch data $d$ in $D$}	    
    	        \STATE Prediction $pred \leftarrow M(d)$
    	        \STATE Update prediction sequence $S$ of instances in $d$
    	        \STATE Compute loss
    	        \STATE Loss back propagation and optimize $M$
            \ENDFOR
        \ENDFOR
        \STATE Distribution $F_C \leftarrow$ calculate metric $C$ of each sample
        \STATE Fit weibull mixture model to $F_C$
        \STATE Obtain selection threshold $\tau = \alpha_2$
        \STATE $D^{'} = \{(x_i, y_i) | C_i < \tau \}_{i=1}^{|D|}$
		\ENSURE training set $D^{'}$ and Model $M$ for next round
	\end{algorithmic}  
\end{algorithm}

\subsection{Utilizing Data with Incorrect Labels}\label{subsec:semisup}
Note that our sample selection approach discards the identified falsely-labeled data and uses the selected clean data for the model training.
Our method can also be extended and combined with existing semi-supervised methods, where we regard the corrupted data as unlabeled data.
Concretely, we first apply our sample selection framework on the noisy training set to separate it into clean data and corrupted data.
Then we utilize a semi-supervised learning strategy on the combination of the clean data and the unlabeled data derived from the corrupted to further improve the generalization of model.
\section{Experiments}
In this section, we conduct extensive experiments to validate the effectiveness of our method. 
Specifically, we evaluate our method on five popularly used datasets including CIFAR-10, CIFAR-100, Mini-ImageNet, Mini-Webvision and CIFAR-N. Due to the page limitation, the results of CIFAR-N dataset which is also the web noise is in the supplementary material.
We also perform a series of ablation studies and analysis to provide more insights for our method.
We first introduce the experiment setup in Sec.~\ref{subsec:exp_setup}, followed by the experimental results on synthetic noise datasets in Sec.~\ref{subsec:exp_results}.
Then we present the experimental results on web noise datasets in Sec.~\ref{subsec:webnoise}.
Finally, we provide the ablation studies and analysis of our method in Sec.~\ref{subsec:ablation_study}.



\subsection{Experiment Setup}\label{subsec:exp_setup}

\paragraph{Datasets}
To systematically evaluate the noisy label methods, we conduct exhaustive experiments on both synthetic and web noise datasets.
For synthetic noise, we follow the protocol proposed in ~\cite{FINE} to generate symmetric and asymmetric noise on CIFAR-10~\cite{cifar-dataset} and CIFAR-100~\cite{cifar-dataset} datasets.
Symmetric noise is provided by randomly flipping the label of a fraction of samples.
In order to make asymmetric noise, sample labels are changed to a specified other class.
Specifically, the mapping relationship for asymmetric noise in CIFAR-10 is as follows: truck $\to$ car, bird $\to$ airplane, deer $\to$ horse, cat $\to$ dog and dog $\to$ cat. For CIFAR-100 dataset, the asymmetric noise is generated by circularly flipping each class to the next class inside super-classes. For real scenarios with web noises, We adopt Mini-Imagenet, Mini-Webvision and CIFAR-N datasets for evaluation.
Besides, Mini-ImageNet~\cite{mini-imagenet} collects noisy images from Internet. Specifically, it annotates the correctness of the collected image labels manually. The dataset includes around 50,000 training images and 5,000 val images for 100 classes.
Mini-Webvision~\cite{webvision} is derived from Webvision dataset~\cite{webvision} by picking top 50 categories of google images.
Note that Webvision dataset is built through keyword searches on Flickr and Google Images, with the 1000 ImageNet categories selected as keywords.
Mini-Webvision dataset consists of 65,944 images for training and 2500 images for test.
CIFAR-N dataset involves realistic annotation noise and more details are in supplementary material.

\begin{table*}[t]
    \renewcommand\arraystretch{1.2}
    \centering
    \begin{tabular}{c|c|ccc} 
    \toprule
    \multirow{2}{*}{Methods} & \multirow{2}{*}{Semi-Tech} & \multicolumn{3}{c}{Noise Ratio}  \\ 
                             &                            & 20\%  & 40\%  & 80\%             \\ 
    \midrule
    \midrule
    Mix\cite{mix}                      & mixup                      & 54.60 & 50.40 & 37.32            \\
    DivideMix\cite{li2020dividemix}                & mixmatch                   & 50.30 & 50.94 & 35.42            \\
    ELR\cite{elr}                      & temporal ensembling                       & 58.10 & 50.62 & 41.68            \\
    MOiT\cite{moit}$^\dag$ & mixup + contrastive learning       & 61.62 & 58.10 & 43.84            \\
    \midrule
    MILD-MM             & mixmatch                   & \textbf{62.56}  & \textbf{59.14}  & \textbf{44.78}             \\
    MILD-MUCL                  & mixup + contrastive learning       & 62.46 & 59.12 & 44.68            \\
    \bottomrule
    \end{tabular}
    \caption{Test accuracy (\%) at final epoch  on Mini-Imagenet dataset. $\dag$ denotes the results are obtained by running their code.}
    \label{mini-imagenet-result}
\end{table*}

\begin{table*}[t]
    \renewcommand\arraystretch{1.2}
    \centering
    \label{mini-webvision-result}
    {\resizebox{\linewidth}{!}{
    \begin{tabular}{c|c|c|c|c|c} 
    \toprule
    Methods      & Semi-Tech            & Backbone                             & Accuracy & Backbone                  & Accuracy        \\ 
    \midrule
    \midrule
    Co-teaching\cite{coteaching}  & -                    & \multirow{2}{*}{Inception-resnet v2} & 63.58    & \multirow{2}{*}{Resnet18} & -               \\
    FINE\cite{FINE}         & -                    &                                      & 75.24    &                           & -               \\
    \midrule
    MILD          & -                    & Inception-resnet v2 & \textbf{76.63}    & Resnet18 & \textbf{73.42}           \\ 
    \midrule
    \midrule
    Mix\cite{mix}          & mixup                & \multirow{4}{*}{Inception-resnet v2} & -        & \multirow{4}{*}{Resnet18} & 73.76           \\
    ELR\cite{elr}          & temporal ensembling                 &                                      & 76.26    &                           & 71.88           \\
    DivideMix\cite{li2020dividemix}    & mixmatch             &                                      & 77.32    &                           & 74.64           \\
    MOiT\cite{moit}         & mixup + contrastive learning &                                      & 80.00$^\dag$    &                           & 77.76           \\
    \midrule
    MILD-MM & mixmatch             &  \multirow{2}{*}{Inception-resnet v2}    & 78.88    & \multirow{2}{*}{Resnet18} & 78.08   \\
    MILD-MUCL      & mixup + contrastive learning &                                      & \textbf{80.52}    &                           & \textbf{78.52}  \\
    \bottomrule
    \end{tabular}}}
    \caption{Test accuracy (\%) at final epoch on Mini-Webvision. $\dag$ denotes the results are obtained by running their code.}
    \label{mini-webvision-result}
\end{table*}

\paragraph{Comparison methods}
To demonstrate the effectiveness of our method, we compare our approach with existing sample selection method like FINE~\cite{FINE}, JoCoR~\cite{jocor} and semi-supervised based approaches like DivideMix~\cite{li2020dividemix} and MOIT~\cite{moit}.
It is worth noting that for fair comparison with semi-supervised based approaches, we expand our approach by employing the same semi-supervised techniques.
Concretely, MixMatch~\cite{berthelot2019mixmatch} used in DivideMix is applied on our approach, named as MILD-MM.
MixUp~\cite{zhang2017mixup} and contrastive learning used in MOiT are introduced into our approach, named as MILD-MUCL.
Note that we take the same configurations with MOiT and DivideMix for the hyper-parameters about semi-supervised.

\paragraph{Implementation details}\label{paragraph:implementation_details}

Our implementation is based on PyTorch.
For CIFAR-10 and CIFAR-100 dataset, the image size is 32x32.
We utilize the ResNet-34 for CIFAR-10 and CIFAR-100 and adopt the same data augmentations as FINE\cite{FINE}, including random cropping and random horizontal flipping. 
We adopt the SGD optimizer with an initial learning rate of 0.01, a momentum of 0.9. 
For Mini-Webvision and Mini-ImageNet, the image size is resized to 224x224 and 84x84, respectively.
We conduct experiments using both Inception-ResNet v2~\cite{inception} and 18-layers ResNet~\cite{resnet} on Mini-Webvision dataset following~\cite{li2020dividemix}.
Besides, 18-layers ResNet is used for Mini-Imagenet following~\cite{moit}. 
We use the SGD optimizer with an initial learning rate of 0.1, a momentum of 0.9.
For all datasets, we use the cosine-annealing learning rate scheduler for each round and the batch size is set to 128.
The training epochs of each round is 20, 50, 100, 100 for CIFAR-10, CIFAR-100, Mini-Imagenet and Mini-Webvision, respectively.
Following FINE~\cite{FINE}, we tune the hyper-parameters by an additional validation set for CIFAR-10 and CIFAR-100.
Also following DivideMix~\cite{li2020dividemix}, the validation set is provided to tune the hyper-parameters for web noise datasets like Mini-Webvision and Mini-ImageNet.



\subsection{Experiments on Synthetic Noise Datasets}\label{subsec:exp_results}

For synthetic noise datasets like CIFAR-10 and CIFAR-100, we first compare our approach with sample selection approaches, shown in Table.~\ref{cifar-results}.
We can see that our method consistently outperforms other selection methods under different types and noise rates on both datasets.
It is worth noting that in general, the performance gain produced by our method improves as the noise rate increases. 
Specifically, compared to FINE, we improve the accuracy from $25.6$ to $36.0$(\textbf{+10.4\%}) under the symmetric 80\% noise of CIFAR-100 dataset.



\subsection{Experiments on Web Noise Datasets}\label{subsec:webnoise}
For web noise dataset, we conduct extensive experiments on Mini-ImageNet and Mini-Webvision.
For Mini-ImageNet dataset, we compare our approach with semi-supervised based methods in Table~\ref{mini-imagenet-result}.
It is shown that our method consistently surpasses other methods under different noise ratios in real noise scenario.
Specifically, MILD-MUCL surpasses MOiT with 0.84$\%$ under 20\% noise ratio, and MILD-MM outperforms than DivideMix with 9.36\% accuracy on 40\% noise ratio.

For Mini-Webvision dataset, we compare both sample-selection and semi-supervised based approaches in Table~\ref{mini-webvision-result}.
Specifically, we compare two different backbones including ResNet18 and Inception-ResNet v2.
It is evident that our method consistently performs better than other sample-selection methods.
Specifically, we surpass FINE with 1.4$\%$ performance improvement on Inception-ResNet v2 backbone.
Besides, using the same semi-supervised technique, our method consistently outperforms than other semi-supervised methods.
Concretely, our method(MILD-MM) surpasses DivideMix with 1.56\% performance gain using Inception-ResNet v2 backbone and 3.44\% improvement on ResNet18 backbone, demonstrating the robustness of our technique to the model architecture.
Compared to MOiT, MILD-MUCL achieves 0.52$\%$ improvement with ResNet18 backbone, which demonstrates the superiority of our proposed approach.
In summary, we can see that our approach can benefit from kinds of semi-supervised techniques.

\begin{figure}[t] 
\centering 
\includegraphics[width=\linewidth]{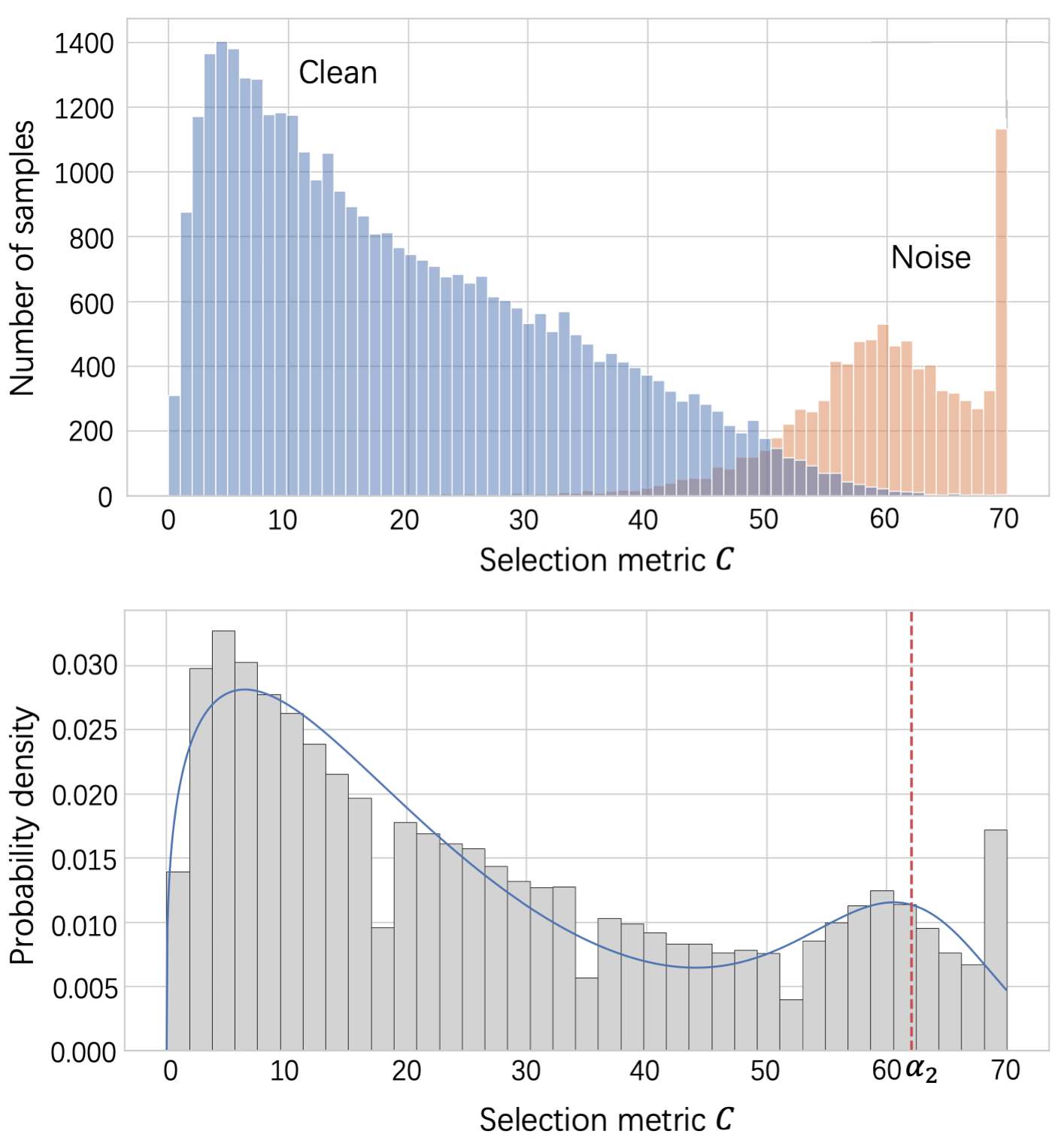} 
\caption{The visualization of metric distribution and mixture model on CIFAR-100 symmetric 20\% noise.} 
\vspace{-2mm}
\label{vis} 

\end{figure}


\begin{table*}[t]
    \renewcommand\arraystretch{1.1}
    \centering
    \begin{tabular}{ccc|ccc|ccc}
    \toprule
    \multirow{2}{*}{Memorization} & \multirow{2}{*}{Forgetting} & \multirow{2}{*}{Mixture Model} & \multicolumn{3}{c|}{CIFAR-10} & \multicolumn{3}{c}{CIFAR-100} \\
                                  &                             &                                & Precision & Recall & Accuracy & Precision & Recall & Accuracy \\ \midrule
    \checkmark                           &                             &                                & 86.75     & 70.91  & 72.76    & 77.85     & 36.92  & 30.37    \\
    \checkmark                           & \checkmark                         &                                & 92.73     & \textbf{75.06}  & 77.56    & 79.61     & \textbf{37.79}  & 33.29    \\
    \checkmark                           & \checkmark                         & \checkmark                            & \textbf{93.37}     & 71.85  & \textbf{79.11}    & \textbf{85.41}     & 37.17  & \textbf{36.03}    \\ \bottomrule
    \end{tabular}
    \caption{Components ablation study on symmetric 80\% noise.}
    \label{memorization-forgetting}
\end{table*}

\begin{table*}[t]
    \centering
    \renewcommand\arraystretch{1.1}
    \begin{tabular}{c|ccc|ccc} 
    \toprule
    \multirow{2}{*}{metric} & \multicolumn{3}{c|}{CIFAR-10} & \multicolumn{3}{c}{CIFAR-100}  \\
                            & Precision & Recall & Accuracy           & Precision & Recall & Accuracy           \\ 
    \midrule
    Small-loss\cite{shen2019learning}      & 97.75      & 95.31   & 92.30             & 88.37      & 88.15   & 67.79               \\
    Eigenvector\cite{FINE}         & \textbf{99.91}     & 77.05  & 91.04              & \textbf{99.56}     & 82.43  & 70.39              \\
    MILD          & 98.35     & \textbf{98.96}  & \textbf{93.03}              & 95.74     & \textbf{98.78}  & \textbf{74.17}              \\
    \bottomrule
    \end{tabular}
    \caption{Performance (\%) comparison with other sample selection methods on symmetric 20\% noise.}
    \label{compare-other-sample-selection}
\end{table*}




\subsection{Ablation Study and Analysis}\label{subsec:ablation_study}

\paragraph{Visualization.} Figure.~\ref{vis} shows the histogram of metric values for clean data(blue columns) and falsely-labeled(orange columns) data and the fitting result of mixture model under 20\% symmetric noise on CIFAR-100 dataset.
We can see that with our proposed metric and mixture model, the clean data and falsely-labeled data can be effectively separated, which can be further utilized to remove falsely-labeled data to alleviate the influence of noises on subsequent training.


\paragraph{Component analysis.}\label{ablation:compoent-analysis}
To provided more insights on the effectiveness of the components in our framework, we conduct ablation experiments in Table.\ref{memorization-forgetting}.
We use another simple selection strategy to compare with the mixture model. Based on the metric value $C_i$, we select $R|D|$ instances with smallest $C_i$ as the training set for subsequent round, where the selection ratio $R$ is set to 0.9.
As Table.\ref{memorization-forgetting} shows, it is indicated that combining memorization and forgetting can improve the selection quality, which highlights the importance of considering overall learning dynamics.
Meanwhile, we can see that the mixture model can further improve the performance, which indicates that the mixture model can benefit the sample selection.

\paragraph{Comparison with sample selection metrics.} We also conduct the metric comparison with small-loss and eigenvector-based\cite{FINE} in same sample selection framework.
For example, we perform multi-round learning with small loss.
Note that in each round, different from existing small-loss methods, we select the best epoch with highest test accuracy for small loss selection.
Table.\ref{compare-other-sample-selection} compares the precision, recall of the selection data and test accuracy for different methods on 20\% symmetric noise of CIFAR datasets. 
We can see that our method achieves better recall when keeping the comparable precision with Eigenvector.

\vspace{-2mm}

\section{Conclusion}\label{section:conclusion}
Deep networks easily memorize the patterns of noisy data, which causes performance degradation.
To alleviate the influence of falsely-labeled data, it is critical to identify correctly labeled data from a noisy training dataset.
In this work, we propose a novel sample selection framework to differentiate clean and corrupted data via their learning dynamics.
We measure the difficulty of memorization and forgetting to represent learning dynamics.
Concretely, the difficulty of memorization is estimated with the transition times from misclassified to memorized.
Besides, the difficulty of forgetting is estimated with the transition times from memorized to misclassified.
We combine them to build the selection criterion as memorization and forgetting are complementary to identify clean data.
We alternating the model training and selecting clean data to keep improving the quality of selected dataset. 
To verify the effectiveness of proposed method, we conduct experiments on synthetic noise datasets  and web noise datasets.
The experimental results show that our method consistently outperforms than other comparison methods.
Despite the fact that our method consistently beats other methods, it is worth noting that there is still a considerable room for improvement given the  recall of our method at extremely high noise rate, such as $80\%$.
For negative social impact, it is possible to be leveraged by malicious applications for our method to improve their application.

\section*{Acknowledgments}

This work was supported by Shanghai Science and Technology Program 21010502700 and Shanghai Frontiers Science Center of Human-centered Artificial Intelligence.

\bibliographystyle{named}
\bibliography{ijcai23}

\begin{thebibliography}{}

\bibitem[\protect\citeauthoryear{Arazo \bgroup \em et al.\egroup
  }{2019}]{arazo2019unsupervised}
Eric Arazo, Diego Ortego, Paul Albert, Noel O’Connor, and Kevin McGuinness.
\newblock Unsupervised label noise modeling and loss correction.
\newblock In {\em International conference on machine learning(ICML)}, 2019.

\bibitem[\protect\citeauthoryear{Arpit \bgroup \em et al.\egroup
  }{2017}]{arpit2017closer}
Devansh Arpit, Stanis{\l}aw Jastrzebski, Nicolas Ballas, David Krueger,
  Emmanuel Bengio, Maxinder~S Kanwal, Tegan Maharaj, Asja Fischer, Aaron
  Courville, Yoshua Bengio, et~al.
\newblock A closer look at memorization in deep networks.
\newblock In {\em International conference on machine learning(ICML)}, 2017.

\bibitem[\protect\citeauthoryear{Berthelot \bgroup \em et al.\egroup
  }{2019}]{berthelot2019mixmatch}
David Berthelot, Nicholas Carlini, Ian Goodfellow, Nicolas Papernot, Avital
  Oliver, and Colin~A Raffel.
\newblock Mixmatch: A holistic approach to semi-supervised learning.
\newblock {\em Advances in Neural Information Processing Systems}, 2019.

\bibitem[\protect\citeauthoryear{Blum \bgroup \em et al.\egroup
  }{2003}]{blum2003noise}
Avrim Blum, Adam Kalai, and Hal Wasserman.
\newblock Noise-tolerant learning, the parity problem, and the statistical
  query model.
\newblock {\em Journal of the ACM (JACM)}, 2003.

\bibitem[\protect\citeauthoryear{Chen \bgroup \em et al.\egroup
  }{2019}]{chen2019understanding}
Pengfei Chen, Ben~Ben Liao, Guangyong Chen, and Shengyu Zhang.
\newblock Understanding and utilizing deep neural networks trained with noisy
  labels.
\newblock In {\em International Conference on Machine Learning(ICML)}, 2019.

\bibitem[\protect\citeauthoryear{Cheng \bgroup \em et al.\egroup
  }{2020}]{CORES}
Hao Cheng, Zhaowei Zhu, Xingyu Li, Yifei Gong, Xing Sun, and Yang Liu.
\newblock Learning with instance-dependent label noise: A sample sieve
  approach.
\newblock {\em arXiv preprint arXiv:2010.02347}, 2020.

\bibitem[\protect\citeauthoryear{Feng \bgroup \em et al.\egroup
  }{2020}]{feng2020deep}
Di~Feng, Christian Haase-Sch{\"u}tz, Lars Rosenbaum, Heinz Hertlein, Claudius
  Glaeser, Fabian Timm, Werner Wiesbeck, and Klaus Dietmayer.
\newblock Deep multi-modal object detection and semantic segmentation for
  autonomous driving: Datasets, methods, and challenges.
\newblock {\em IEEE Transactions on Intelligent Transportation Systems}, 2020.

\bibitem[\protect\citeauthoryear{Ghosh \bgroup \em et al.\egroup
  }{2017}]{ghosh2017robust}
Aritra Ghosh, Himanshu Kumar, and PS~Sastry.
\newblock Robust loss functions under label noise for deep neural networks.
\newblock In {\em Proceedings of the AAAI conference on artificial
  intelligence(AAAI)}, 2017.

\bibitem[\protect\citeauthoryear{Han \bgroup \em et al.\egroup
  }{2018a}]{han2018co}
Bo~Han, Quanming Yao, Xingrui Yu, Gang Niu, Miao Xu, Weihua Hu, Ivor Tsang, and
  Masashi Sugiyama.
\newblock Co-teaching: Robust training of deep neural networks with extremely
  noisy labels.
\newblock {\em Advances in neural information processing systems(NeurIPS},
  2018.

\bibitem[\protect\citeauthoryear{Han \bgroup \em et al.\egroup
  }{2018b}]{coteaching}
Bo~Han, Quanming Yao, Xingrui Yu, Gang Niu, Miao Xu, Weihua Hu, Ivor Tsang, and
  Masashi Sugiyama.
\newblock Co-teaching: Robust training of deep neural networks with extremely
  noisy labels.
\newblock {\em Advances in neural information processing systems}, 31, 2018.

\bibitem[\protect\citeauthoryear{He \bgroup \em et al.\egroup }{2016}]{resnet}
Kaiming He, Xiangyu Zhang, Shaoqing Ren, and Jian Sun.
\newblock Deep residual learning for image recognition.
\newblock In {\em Proceedings of the IEEE conference on computer vision and
  pattern recognition}, pages 770--778, 2016.

\bibitem[\protect\citeauthoryear{Jiang \bgroup \em et al.\egroup
  }{2018}]{jiang2018mentornet}
Lu~Jiang, Zhengyuan Zhou, Thomas Leung, Li-Jia Li, and Li~Fei-Fei.
\newblock Mentornet: Learning data-driven curriculum for very deep neural
  networks on corrupted labels.
\newblock In {\em International Conference on Machine Learning(ICML)}, 2018.

\bibitem[\protect\citeauthoryear{Jiang \bgroup \em et al.\egroup
  }{2020}]{mini-imagenet}
Lu~Jiang, Di~Huang, Mason Liu, and Weilong Yang.
\newblock Beyond synthetic noise: Deep learning on controlled noisy labels.
\newblock In {\em International Conference on Machine Learning(ICML)}, 2020.

\bibitem[\protect\citeauthoryear{Kim \bgroup \em et al.\egroup }{2021}]{FINE}
Taehyeon Kim, Jongwoo Ko, JinHwan Choi, Se-Young Yun, et~al.
\newblock Fine samples for learning with noisy labels.
\newblock {\em Advances in Neural Information Processing Systems}, 34, 2021.

\bibitem[\protect\citeauthoryear{Krizhevsky \bgroup \em et al.\egroup
  }{2009}]{cifar-dataset}
Alex Krizhevsky, Geoffrey Hinton, et~al.
\newblock Learning multiple layers of features from tiny images.
\newblock {\em Technical report, University of Toronto}, 2009.

\bibitem[\protect\citeauthoryear{Li \bgroup \em et al.\egroup
  }{2017}]{webvision}
Wen Li, Limin Wang, Wei Li, Eirikur Agustsson, and Luc Van~Gool.
\newblock Webvision database: Visual learning and understanding from web data.
\newblock {\em arXiv preprint arXiv:1708.02862}, 2017.

\bibitem[\protect\citeauthoryear{Li \bgroup \em et al.\egroup
  }{2020}]{li2020dividemix}
Junnan Li, Richard Socher, and Steven~CH Hoi.
\newblock Dividemix: Learning with noisy labels as semi-supervised learning.
\newblock {\em arXiv preprint arXiv:2002.07394}, 2020.

\bibitem[\protect\citeauthoryear{Li \bgroup \em et al.\egroup
  }{2021}]{li2021superpixel}
Shuailin Li, Zhitong Gao, and Xuming He.
\newblock Superpixel-guided iterative learning from noisy labels for medical
  image segmentation.
\newblock In {\em International Conference on Medical Image Computing and
  Computer-Assisted Intervention(MICCAI)}, 2021.

\bibitem[\protect\citeauthoryear{Liu \bgroup \em et al.\egroup }{2020}]{elr}
Sheng Liu, Jonathan Niles-Weed, Narges Razavian, and Carlos Fernandez-Granda.
\newblock Early-learning regularization prevents memorization of noisy labels.
\newblock {\em Advances in neural information processing systems},
  33:20331--20342, 2020.

\bibitem[\protect\citeauthoryear{Ma \bgroup \em et al.\egroup
  }{2020}]{ma2020normalized}
Xingjun Ma, Hanxun Huang, Yisen Wang, Simone Romano, Sarah Erfani, and James
  Bailey.
\newblock Normalized loss functions for deep learning with noisy labels.
\newblock In {\em International Conference on Machine Learning(ICML)}, 2020.

\bibitem[\protect\citeauthoryear{Mirzasoleiman \bgroup \em et al.\egroup
  }{2020}]{crust}
Baharan Mirzasoleiman, Kaidi Cao, and Jure Leskovec.
\newblock Coresets for robust training of neural networks against noisy labels.
\newblock {\em arXiv preprint arXiv:2011.07451}, 2020.

\bibitem[\protect\citeauthoryear{Nguyen \bgroup \em et al.\egroup
  }{2019}]{nguyen2019self}
Duc~Tam Nguyen, Chaithanya~Kumar Mummadi, Thi Phuong~Nhung Ngo, Thi Hoai~Phuong
  Nguyen, Laura Beggel, and Thomas Brox.
\newblock Self: Learning to filter noisy labels with self-ensembling.
\newblock {\em arXiv preprint arXiv:1910.01842}, 2019.

\bibitem[\protect\citeauthoryear{Ortego \bgroup \em et al.\egroup
  }{2021}]{moit}
Diego Ortego, Eric Arazo, Paul Albert, Noel~E O'Connor, and Kevin McGuinness.
\newblock Multi-objective interpolation training for robustness to label noise.
\newblock In {\em Proceedings of the IEEE/CVF Conference on Computer Vision and
  Pattern Recognition}, pages 6606--6615, 2021.

\bibitem[\protect\citeauthoryear{Patrini \bgroup \em et al.\egroup
  }{2017}]{forward}
Giorgio Patrini, Alessandro Rozza, Aditya Krishna~Menon, Richard Nock, and
  Lizhen Qu.
\newblock Making deep neural networks robust to label noise: A loss correction
  approach.
\newblock In {\em Proceedings of the IEEE conference on computer vision and
  pattern recognition}, pages 1944--1952, 2017.

\bibitem[\protect\citeauthoryear{Reed \bgroup \em et al.\egroup
  }{2014}]{bootstrap}
Scott Reed, Honglak Lee, Dragomir Anguelov, Christian Szegedy, Dumitru Erhan,
  and Andrew Rabinovich.
\newblock Training deep neural networks on noisy labels with bootstrapping.
\newblock {\em arXiv preprint arXiv:1412.6596}, 2014.

\bibitem[\protect\citeauthoryear{Shen and Sanghavi}{2019}]{shen2019learning}
Yanyao Shen and Sujay Sanghavi.
\newblock Learning with bad training data via iterative trimmed loss
  minimization.
\newblock In {\em International Conference on Machine Learning(ICML)}, 2019.

\bibitem[\protect\citeauthoryear{Song \bgroup \em et al.\egroup
  }{2019}]{song2019selfie}
Hwanjun Song, Minseok Kim, and Jae-Gil Lee.
\newblock Selfie: Refurbishing unclean samples for robust deep learning.
\newblock In {\em International Conference on Machine Learning(ICML)}, 2019.

\bibitem[\protect\citeauthoryear{Song \bgroup \em et al.\egroup
  }{2022}]{song2022learning}
Hwanjun Song, Minseok Kim, Dongmin Park, Yooju Shin, and Jae-Gil Lee.
\newblock Learning from noisy labels with deep neural networks: A survey.
\newblock {\em IEEE Transactions on Neural Networks and Learning
  Systems(TNNLS)}, 2022.

\bibitem[\protect\citeauthoryear{Szegedy \bgroup \em et al.\egroup
  }{2017}]{inception}
Christian Szegedy, Sergey Ioffe, Vincent Vanhoucke, and Alexander~A Alemi.
\newblock Inception-v4, inception-resnet and the impact of residual connections
  on learning.
\newblock In {\em Thirty-first AAAI conference on artificial intelligence},
  2017.

\bibitem[\protect\citeauthoryear{Toneva \bgroup \em et al.\egroup
  }{2018}]{toneva2018empirical}
Mariya Toneva, Alessandro Sordoni, Remi~Tachet des Combes, Adam Trischler,
  Yoshua Bengio, and Geoffrey~J Gordon.
\newblock An empirical study of example forgetting during deep neural network
  learning.
\newblock In {\em International Conference on Learning Representations(ICLR)},
  2018.

\bibitem[\protect\citeauthoryear{Wei \bgroup \em et al.\egroup }{2020}]{jocor}
Hongxin Wei, Lei Feng, Xiangyu Chen, and Bo~An.
\newblock Combating noisy labels by agreement: A joint training method with
  co-regularization.
\newblock In {\em Proceedings of the IEEE/CVF Conference on Computer Vision and
  Pattern Recognition}, pages 13726--13735, 2020.

\bibitem[\protect\citeauthoryear{Wei \bgroup \em et al.\egroup
  }{2022}]{CIFARN-dataset}
Jiaheng Wei, Zhaowei Zhu, Hao Cheng, Tongliang Liu, Gang Niu, and Yang Liu.
\newblock Learning with noisy labels revisited: A study using real-world human
  annotations.
\newblock In {\em International Conference on Learning Representations}, 2022.

\bibitem[\protect\citeauthoryear{Wu \bgroup \em et al.\egroup
  }{2020}]{topfilter}
Pengxiang Wu, Songzhu Zheng, Mayank Goswami, Dimitris Metaxas, and Chao Chen.
\newblock A topological filter for learning with label noise.
\newblock {\em Advances in neural information processing systems},
  33:21382--21393, 2020.

\bibitem[\protect\citeauthoryear{Yan \bgroup \em et al.\egroup
  }{2014}]{yan2014learning}
Yan Yan, R{\'o}mer Rosales, Glenn Fung, Ramanathan Subramanian, and Jennifer
  Dy.
\newblock Learning from multiple annotators with varying expertise.
\newblock {\em Machine learning}, 2014.

\bibitem[\protect\citeauthoryear{Yu \bgroup \em et al.\egroup
  }{2019}]{coteaching+}
Xingrui Yu, Bo~Han, Jiangchao Yao, Gang Niu, Ivor Tsang, and Masashi Sugiyama.
\newblock How does disagreement help generalization against label corruption?
\newblock In {\em International Conference on Machine Learning}, pages
  7164--7173. PMLR, 2019.

\bibitem[\protect\citeauthoryear{Zhang and
  Sabuncu}{2018}]{zhang2018generalized}
Zhilu Zhang and Mert Sabuncu.
\newblock Generalized cross entropy loss for training deep neural networks with
  noisy labels.
\newblock {\em Advances in neural information processing systems(NeurIPS)},
  2018.

\bibitem[\protect\citeauthoryear{Zhang \bgroup \em et al.\egroup
  }{2017a}]{zhang2017mixup}
Hongyi Zhang, Moustapha Cisse, Yann~N Dauphin, and David Lopez-Paz.
\newblock mixup: Beyond empirical risk minimization.
\newblock {\em arXiv preprint arXiv:1710.09412}, 2017.

\bibitem[\protect\citeauthoryear{Zhang \bgroup \em et al.\egroup }{2017b}]{mix}
Hongyi Zhang, Moustapha Cisse, Yann~N Dauphin, and David Lopez-Paz.
\newblock mixup: Beyond empirical risk minimization.
\newblock {\em arXiv preprint arXiv:1710.09412}, 2017.

\bibitem[\protect\citeauthoryear{Zhou \bgroup \em et al.\egroup
  }{2020}]{zhou2020robust}
Tianyi Zhou, Shengjie Wang, and Jeff Bilmes.
\newblock Robust curriculum learning: from clean label detection to noisy label
  self-correction.
\newblock In {\em International Conference on Learning Representations}, 2020.

\bibitem[\protect\citeauthoryear{Zhu \bgroup \em et al.\egroup
  }{2021a}]{zhu2021understanding}
Jianing Zhu, Jingfeng Zhang, Bo~Han, Tongliang Liu, Gang Niu, Hongxia Yang,
  Mohan Kankanhalli, and Masashi Sugiyama.
\newblock Understanding the interaction of adversarial training with noisy
  labels.
\newblock {\em arXiv preprint arXiv:2102.03482}, 2021.

\bibitem[\protect\citeauthoryear{Zhu \bgroup \em et al.\egroup }{2021b}]{CAL}
Zhaowei Zhu, Tongliang Liu, and Yang Liu.
\newblock A second-order approach to learning with instance-dependent label
  noise.
\newblock In {\em Proceedings of the IEEE/CVF Conference on Computer Vision and
  Pattern Recognition}, pages 10113--10123, 2021.

\end{thebibliography}

\clearpage
\setcounter{section}{0}
\renewcommand\thesection{\Alph{section}}

\begin{table*}[h]
    \renewcommand\arraystretch{1.5}
    \centering
    \caption{Test accuracy (\%) on CIFAR-10N and CIFAR-100N}
    \begin{tabular}{c|ccccc|c}
    \toprule
    \multirow{2}{*}{Method} & \multicolumn{5}{c|}{CIFAR10-N}                                                                               & CIFAR100-N           \\
                            & Aggregate           & Random1             & Random2             & Random3            & Worst               & Noisy Fine          \\ \midrule
    Standard                & 87.77 $\pm$ 0.38          & 85.02 $\pm$ 0.65          & 86.46 $\pm$ 1.79          & 85.16 $\pm$ 0.61          & 77.69 $\pm$ 1.55          & 55.50 $\pm$ 0.66          \\
    Co-Teaching\cite{coteaching}             & 91.20 $\pm$ 0.13          & 90.33 $\pm$ 0.13          & 90.30 $\pm$ 0.17          & 90.15 $\pm$ 0.18          & 83.83 $\pm$ 0.13          & 60.37 $\pm$ 0.27          \\
    JoCoR\cite{jocor}                   & 91.44 $\pm$ 0.05          & 90.30 $\pm$ 0.20          & 90.21 $\pm$ 0.19          & 90.11 $\pm$ 0.21          & 83.37 $\pm$ 0.30          & 59.97 $\pm$ 0.24          \\
    ELR\cite{elr}                     & 92.38 $\pm$ 0.64          & 91.46 $\pm$ 0.38          & 91.61 $\pm$ 0.16          & 91.41 $\pm$ 0.44          & 83.58 $\pm$ 1.13          & 58.94 $\pm$ 0.92          \\
    CAL\cite{CAL}                     & 91.97 $\pm$ 0.32          & 90.93 $\pm$ 0.31          & 90.75 $\pm$ 0.30          & 90.74 $\pm$ 0.24          & 85.36 $\pm$ 0.16          & 61.73 $\pm$ 0.42          \\
    CORES\cite{CORES}                   & 91.23 $\pm$ 0.11          & 89.66 $\pm$ 0.32          & 89.91 $\pm$ 0.45          & 89.79 $\pm$ 0.50          & 83.60 $\pm$ 0.53          & 61.15 $\pm$ 0.73          \\
    MILD                    & \textbf{93.18 $\pm$ 0.23} & \textbf{91.74 $\pm$ 0.15} & \textbf{91.96 $\pm$ 0.13} & \textbf{91.49 $\pm$ 0.11} & \textbf{87.32 $\pm$ 0.58} & \textbf{65.10 $\pm$ 0.67} \\ \bottomrule
    \end{tabular}
    \label{cifarn}
\end{table*}

\begin{table*}[t]
    \centering
    \renewcommand\arraystretch{1.5}
    \caption{Precision, recall and accuracy (\%) of each round on CIFAR-100 symmetric 80\% noise}
    \begin{tabular}{c|cccccccc} 
    \toprule
    Round & 1     & 2     & 3     & 4    & 5             & 6    & 7 \\ 
    \midrule
    Precision & 20.86 & 27.76 & 36.47 & 46.63 & 60.01 & 66.64 & 85.41 \\
    Recall    & 100.0 & 87.45 & 80.60 & 74.17 & 59.68 & 55.52 & 37.17 \\
    Accuracy  & 9.38  & 13.62 & 19.39 & 25.68 & 30.47 & 32.95 & 36.03 \\
    \bottomrule
    \end{tabular}
    \label{iteration-sensitive}
\end{table*}

\begin{table*}[t]
    \renewcommand\arraystretch{1.5}
    \centering
    \caption{Ablation study of metric simplification on symmetric 80\% noise. MILD-F represents the full version of MILD metric. MILD represents the simplified version. }
    \begin{tabular}{c|ccc|ccc} 
    \toprule
    \multirow{2}{*}{Method} & \multicolumn{3}{c|}{CIFAR-10} & \multicolumn{3}{c}{CIFAR-100}  \\ 
                            & Precision & Recall & Accuracy & Precision & Recall & Accuracy  \\ 
    \midrule
    MILD-F                      & 92.73     & 75.06  & 77.56    & 79.61     & 37.79  & 33.29     \\
    MILD                     & 91.88     & 74.28  & 77.25    & 84.49     & 39.94  & 35.00     \\
    \bottomrule
    \end{tabular}
    \label{simplification-ablation}
\end{table*}

\section{Implementation Details}

For Inception-ResNet v2 backbone on Mini-Webvision dataset, we follow \cite{li2020dividemix} to apply augmentations consists of random resized cropping and horizontal flipping.
Moreover, for 18-layers ResNet on Mini-Webvision dataset, we follow~ \cite{FINE} to apply augmentations consists of random resized cropping, horizontal flipping, color jittering and random gray scaling.
For Mini-ImageNet, we adopt the augmentations including random resized cropping, horizontal flipping and color jittering.

For CIFAR-10 and CIFAR-100 datasets, we use a single NVIDIA GeForce RTX 2080 Ti GPU. For Mini-ImageNet, we use two NVIDIA GeForce RTX 2080 Ti GPUs. For Mini-Webvision, we use two NVIDIA TITAN RTX GPUs for both Inception-ResNet v2 backbone and 18-layers ResNet.

\section{Experiments on CIFAR-N Dataset}

CIFAR-N~\cite{CIFARN-dataset} is also a web noise dataset, in which the incorrect labels are all annotated by human annotators. We followed \cite{CIFARN-dataset} and conducted experiments under the same settings. From Table.\ref{cifarn} we can see that MILD achieves the best performance under all noise settings. According to the noise ratios given by \cite{CIFARN-dataset} under each noise setting, the noise ratios of random1, random2 and random3 on CIFAR10-N are less than 20\%. Therefore, we slightly outperform other methods under these three noise settings. Moreover, On the more challenging CIFAR100-N dataset with a larger noise ratio(40.20\%), our method outperforms the CAL by \textbf{3.37\%}, illustrating the effectiveness of our approach.

\section{Ablation of Iterative Training}

We investigate the influence of rounds on precision, recall, and accuracy in CIFAR100 dataset with 80\% symmetric noise. As seen in Table.\ref{iteration-sensitive}, as the number of rounds grows, the accuracy increases while the precision increases and the recall decreases.

\section{Ablation of Metric Simplification}

We conduct experiments to compare the full version metric $C^F$ and the simplified version metric $C$. As shown in Table.\ref{simplification-ablation}, the comparable results between MILD-F and MILD illustrate the simplified metric is close to $C^F$, which also integrates memorization and forgetting.

\end{document}